%% file: ms.tex
\pdfoutput=1
\documentclass{article}
\pdfoutput=1

\usepackage[T1]{fontenc}
\usepackage{lmodern}

\usepackage[utf8]{inputenc} 
\usepackage[T1]{fontenc}    
\usepackage{hyperref}       
\usepackage{url}            
\usepackage{booktabs}       
\usepackage{amsfonts}       
\usepackage{nicefrac}       
\usepackage{microtype}      

\usepackage{enumerate}

\usepackage{graphicx}
\usepackage{array}

\def\PORR{PaRR}
\def\SH{Sequential Halving}
\def\DL{Deep Learning}

\title{Toward Optimal Run Racing:\\ Application to Deep Learning Calibration}

\author{
Olivier Bousquet$^1$, Sylvain Gelly$^1$, Karol Kurach$^1$,\\
Marc Schoenauer$^2$, Mich\`ele Sebag$^2$, Olivier Teytaud$^1$,\\
  Damien Vincent$^1$.\\ 1. Google Brain, 2. TAU, Inria Saclay IDF.\\ {\texttt{oteytaud@google.com}} 
}

\begin{document}

\maketitle

\begin{abstract}
This paper aims at one-shot learning of deep neural nets, where a highly parallel setting is considered to address the algorithm calibration problem $-$ selecting the best neural architecture and learning hyper-parameter values depending on the dataset at hand. The notoriously expensive calibration problem is optimally reduced by detecting and early stopping non-optimal runs. The theoretical contribution regards the optimality guarantees within the  multiple hypothesis testing framework. Experimentations on the Cifar10, PTB and Wiki benchmarks demonstrate the relevance of the approach with a principled and consistent improvement on the state of the art \cite{Domhan} with no extra  hyper-parameter.
\end{abstract}

\input{Intro_v3}

\input{Related_v2}

\input{Methods}

\input{Experiments}


\bibliographystyle{abbrv}
\bibliography{mybib,ALORS,neo_bib}

\cleardoublepage

{\bf\large\center{Supplementary material.}}
\appendix

\section{Experiments with other values of $\delta$}
\subsection{Comparison between criteria with $\delta=0.1$}

{\scriptsize{\begin{tabular}{|@{}m{2cm}@{}|@{}m{2cm}@{}|@{}m{2cm}@{}|@{}m{2cm}@{}|@{}m{2cm}@{}|@{}m{2cm}@{}|}
\hline
 Testbed &
 Budget (number of runs) &
 Computational cost saved up by method (a) (equal to Domhan et al) &
 C.c. saved up by method (c): prediction-halt operator. &
 C.c. saved up by method (e): best-prediction-halt operator. &
 C.c. saved up by method ``Clever-halt'' \\
\hline
 Cifar-adagrad  &
 22 &
 -50.7\% &
 -50.7\% &
 \bf -57.7\% &
 -50.7\% \\
\hline
 Cifar-adam &
 22 &
 \bf -87.0\% &
 -92.4\% FAIL by 0.797 $\rightarrow$ 0.923 &
 -95.2\% FAIL by 0.797 $\rightarrow$ 0.933 &
 \bf -87.0\% \\
\hline
 Cifar-gradient &
 22 &
 \bf -79.0\% &
 \bf -79.0\% &
 \bf -79.0\% &
 \bf -79.0\% \\
\hline
 Cifar-momentum &
 22 &
 -81.3\% &
 \bf -85.2\% &
 \bf -85.2\% &
 -81.3\% \\
\hline
 Miniwiki bits &
 250 &
 -30.0\% &
 -30.0\% &
 \bf -35.2\% &
 -30.0\% \\
\hline
 Miniwiki bytes &
 250 &
 \bf -12.0\% &
 \bf -12.0\% &
 \bf -12.0\% &
 \bf -12.0\% \\
\hline
 PTB bits &
 250 &
 -23.6\% &
 -23.6\% &
 \bf -26.0\% &
 -23.6\% \\
\hline
 PTB bytes &
 250 &
 \bf -19.73\% &
 \bf -19.73\% &
 \bf -19.73\% &
 \bf -19.73\% \\
\hline
 PTB words &
 250 &
 -52.09\% &
 -60.16 &
 \bf -63.33 &
 -52.12 \\
\hline
 Miniwiki bits &
 50 &
 \bf -36.3\% &
 \bf -36.3\% &
 \bf -36.3\% &
 \bf -36.3\% \\
\hline
 Miniwiki bytes &
 50 &
 -33.5\% &
 -33.5\% &
 \bf -62.3\% &
 -33.5\% \\
\hline
 PTB bits &
 50 &
 \bf -28.7\% &
 \bf -28.7\% &
 \bf -28.7\% &
 \bf -28.7\% \\
\hline
 PTB bytes &
 50 &
 \bf -19.4\% &
 \bf -19.4\% &
 \bf -19.4\% &
 \bf -19.4\% \\
\hline
 PTB words &
 50 &
 -51.5\% &
 \bf -59.4\% &
 -67\% FAIL by 1.201 $\rightarrow$ 1.238 &
 -52.3\% \\
\hline
   &
    &
     &
      &
       &
      No fail, best mean performance. \\
      \hline 
\end{tabular}}}

\subsection{Comparison between criteria with $\delta=0.3$}
We checked the stability of the method by also checking what is going on at $\delta=0.3$. We still get the best results with our method.

{\scriptsize{\begin{tabular}{|@{}m{2cm}@{}|@{}m{2cm}@{}|@{}m{2cm}@{}|@{}m{2cm}@{}|@{}m{2cm}@{}|@{}m{2cm}@{}|}
\hline
 Testbed &
 Budget (number of runs) &
 Computational cost saved up by method (a) (equal to Domhan et al) &
 C.c. saved up by method (c): prediction-halt operator. &
 C.c. saved up by method (e): best-prediction-halt operator. &
 C.c. saved up by method ``Clever-halt'' \\
\hline
 Cifar-adagrad  &
 22 &
 -56\% &
 -56\% &
 \bf -73\% &
 -56\% \\
\hline
 Cifar-adam &
 22 &
 \bf -87.5\%  &
  -93.7\% FAIL by 0.797 $\rightarrow$ 0.923 &
  -95.4\% FAIL by 0.797 $\rightarrow$ 0.939 &
  \bf -87.5\% \\
\hline
  Cifar-gradient &
  22 &
  \bf -83.0\% &
  \bf -83.0\% &
  -89.2\%  FAIL by 1.175 $\rightarrow$ 1.200 &
  \bf -83.0\% \\
\hline
  Cifar-momentum &
  22 &
  -81.9\% &
  \bf -85.8\% &
  -91.7\% FAIL by 1.034 $\rightarrow$ 1.146 &
  -81.9\% \\
\hline
  Miniwiki bits &
  250 &
  -39.6\% &
  \bf -41.7\% &
  -47.4\% FAIL by 2.011 to 2.053 &
  -39.6\% \\
\hline
  Miniwiki bytes &
  250 &
  \bf -12.9\% &
  \bf -12.9\% &
  \bf -12.9\% &
  \bf -12.9\% \\
\hline
  PTB bits &
  250 &
  \bf -32.4\% &
  FAIL by 1.403 $\rightarrow$ 1.406 &
  FAIL by 1.403 $\rightarrow$ 1.410 &
  \bf -32.4\% \\
\hline
  PTB bytes &
  250 &
  -20\% &
  \bf -20.3\% &
  \bf -20.3\% &
  -20\% \\
\hline
  PTB words &
  250 &
  -63.3\% &
  \bf -68\% &
  -69\% FAIL by 1.184 to 1.197 &
  -63.3\% \\
\hline
  Miniwiki bits &
  50 &
  \bf -38.9\% &
  \bf -38.9\% &
  \bf -38.9\% &
  \bf -38.9\% \\
\hline
  Miniwiki bytes &
  50 &
  \bf -38.6\% &
  \bf -38.6\% &
  -65.7\% FAIL by 1.883 $\rightarrow$ 1.889 &
  \bf -38.6\% \\
\hline
  PTB bits &
  50 &
\bf  -38.7\% &
\bf  -38.7\% &
  \bf-38.7\% &
  \bf-38.7\% \\
\hline
  PTB bytes &
  50 &
  \bf-20.8\% &
  \bf-20.8\% &
  \bf-20.8\% &
  \bf-20.8\% \\
\hline
  PTB words &
  50 &
  -56.3\% &
  -62.6\% &
  -76.7\% FAIL by 1.201 to 1.257 &
  \bf -56.7\% \\
\hline
    &
     &
      &
       &
        &
       Never fails, best mean performance. \\
       \hline 
\end{tabular}}}

\begin{table}[t]
\center
{\scriptsize{\begin{tabular}{|@{}m{2cm}@{}|@{}m{2cm}@{}|@{}m{2cm}@{}|@{}m{2cm}@{}|@{}m{2cm}@{}|@{}m{2cm}@{}|}
\hline
 Testbed &
 Budget (number of runs) &
 Computational cost saved up by method (a) (equal to Domhan et al) &
 C.c. saved up by method (c): prediction-halt operator. &
 C.c. saved up by method (e): best-prediction-halt operator. &
 C.c. saved up by method ``Clever-halt'' \\
\hline
 Cifar-adagrad  &
 22 &
 -50.2\% &
 -50.2\% &
 \bf -57.2\% &
 -50.2\% \\
\hline
 Cifar-adam &
 22 &
 -85.8 \% &
 -85.8\% &
 -94.75\%  FAIL by 0.80 $\rightarrow$ 0.94 &
 \bf -85.9\% \\
\hline
 Cifar-gradient &
 22 &
 \bf -70.75\% &
 \bf -70.75\% &
\bf  -70.75\% &
 \bf -70.75\% \\
\hline
 Cifar-momentum &
 22 &
 -63.3\% &
 \bf -81.1\% &
 \bf -81.1\% &
 -63.3\% \\
\hline
 Miniwiki bits &
 250 &
 -19\% &
 -19\% &
 \bf -20.5\% &
 -19\% \\
\hline
 Miniwiki bytes &
 250 &
 \bf -9.5\% &
 \bf -9.5\% &
 \bf -9.5\% &
 -9.1\% \\
\hline
 PTB bits &
 250 &
 -17.1\% &
 -17.1\% &
 \bf -19.2\% &
 -17.1\% \\
\hline
 PTB bytes &
 250 &
 \bf -17.7\% &
 \bf -17.7\% &
 \bf -17.7\% &
 \bf -17.7\% \\
\hline
 PTB words &
 250 &
 -40.5\% &
 -51.0\% &
 \bf -56.4\% &
 -44.4\% \\
\hline
 Miniwiki bits &
 50 &
 \bf -32.7\% &
 \bf -32.7\% &
 \bf -32.7\% &
 \bf -32.7\% \\
\hline
 Miniwiki bytes &
 50 &
 -30.5\% &
 -30.5\% &
 \bf -57.8\% &
 -32.7\% \\
\hline
 PTB bits &
 50 &
 \bf -27.1\% &
 \bf -27.1\% &
 \bf -27.1\% &
 \bf -27.1\% \\
\hline
 PTB bytes &
 50 &
 \bf -18\% &
 \bf -18\% &
 \bf -18\% &
 \bf -18\% \\
\hline
 PTB words &
 50 &
 -49.6\% &
 \bf -58.1\% &
 -65.7\% FAIL by 1.2014 $\rightarrow$ 1.23329 &
 -50.3\% \\
\hline
   &
    &
     &
      No fail, best mean performance.      &
       &
 \\
      \hline 
\end{tabular}}}
\caption{\label{zerozeroone}Comparative results under same conditions as in Table \ref{withhacks}, except for confidence $\delta=0.01$.}
\end{table}

\end{document}

%% file: Intro_v3.tex
\section{Introduction}
The algorithm selection problem $-$ aimed at selecting {\em a priori} the learning algorithm  best suited to a given dataset,  and the algorithm calibration problem $-$ aimed at identifying the best hyper-parameter setting of an algorithm for the dataset at hand,  have been acknowledged to be key issues since the late 80s \cite{Bradzil00,Bardenet11,thornton2013AutoWeka,Bardenet13,Pfahringer13,Kalousis14}.
Several challenges have been organized to further investigate both algorithm selection and calibration issues in the last few years \cite{AutoML16-ICML,AutoML15-NIPS}. 
\def\DNN{{DL}}

The algorithm selection issue appears settled as of now, at least in the case where sufficient training data is available: deep learning (\DNN) consistently delivers dominant performances in many application domains, and is currently considered to the best learning algorithm in the large data regime  \cite{KrizhevskySH12,Donahue,Deng,Atari,AlphaGo}.
This renders the algorithm calibration an even more critical issue: on the one hand, \DNN\  notoriously requires high computational resources; on the other hand, it involves a structured hyper-parameter space, hindering the approximation of the performance model. 
Automatic algorithm calibration thus is challenged by {manual} algorithm calibration, as noted by \cite{Domhan}. As the experienced practitioner can easily detect and stop the unpromising runs based on their learning curves in the first epochs, she can afford to consider many more hyper-parameter settings. 

How to discard as early as possible runs/solutions that will eventually yield under-optimal results has long and thoroughly been investigated (section \ref{sec:soa}). The early discard decision problem raises two interdependent questions: uncertainty modelling, as the eventual quality of a run result is unknown until the run is achieved; risk control, as one needs guarantees that the run  which would have yielded the best result has not been stopped.

This paper addresses the early discard problem in the context of parallel one-shot deep neural training. Formally, the considered framework, referred to as  parallel one-shot run race (\PORR), allocates all available computational resources at the beginning of the period to train deep neural nets; each core runs with its specific hyper-parameter setting, or configuration, with no communication among the cores. The  goal is to make \DNN\ robust w.r.t. random hazards (e.g. initializations) and bad decisions (e.g. configuration set), eventually delivering the optimal configuration/learned model,  with a minimal computational budget.
The challenge lies in making the stopping decision with little and censored evidence: as all runs are simultaneous, only prior information about the learning curves behavior is available, as in \cite{Domhan}. 
Formally, the \PORR\ problem is a constrained optimization problem: i) the constraint regards the guarantees about eventually delivering the optimal result, i.e. ensuring that the best run lives until the end of the period; ii) the optimization consists of minimizing the computational budget subject to the optimality guarantee, by stopping any run (with no possible resuming of the run, as opposed to \cite{freezethaw}) as early as possible.

The present paper, building upon the current best approach \cite{Domhan}, makes theoretical and empirical contributions. On the theoretical side and with no additional hyper-parameter, a principled approach is used to set the pruning thresholds; furthermore, guarantees are obtained through a principled treatment of the multiple hypothesis testing issue. On the empirical side, experimentations on the Cifar \cite{cifar10}, PTB \cite{PTB}, and MiniWiki \cite{hutter} benchmarks show a consistent improvement compared to  \cite{Domhan}, for each and every hyper-parameter setting.

 This paper is organized as follows. Section \ref{sec:soa} briefly discusses related work. Section \ref{sec:algo} formulates the \PORR\ problem together with different types of statistical risk and associated criteria. The proposed \PORR\ decision maker, with the different variants associated to the risks, are empirically assessed and compared to the state of the art in section \ref{sec:expe}.

%% file: Related_v2.tex
\section{Related work}\label{sec:soa}
\paragraph{Performance modelling.}
In the domain of algorithm selection and calibration, a usual approach is to build a performance model \cite{rice1976algorithm}, predicting the eventual performance of the algorithm based on its hyper-parameter configuration and on the description of the problem instance at hand. In the Machine Learning domain however, in contrast with the SAT and CSP domains, to our best knowledge there does not exist yet an affordable feature set, able to accurately describe a problem instance and to support the prediction of an algorithm performance on this instance. For this reason, algorithm selection and calibration in Machine Learning (see e.g. \cite{Bardenet11,SnoekLA12,AutoWeka}), builds online an instance-dependent performance model, learned using Gaussian Processes \cite{Bardenet11,SnoekLA12}, Random Forests \cite{AutoWeka} or radius-based functions \cite{hord}). In most cases \cite{Bardenet11,SnoekLA12,AutoWeka} the performance model is used along Bayesian Optimization principles \cite{Bayesian-Mockus} to determine the most promising algorithm configuration. In \cite{hord}, coordinate-based optimization reports good results, particularly so in high-dimensional hyper-parameter space.

By construction, the above approaches are intrinsically sequential, making it difficult to use the above performance models to stop unpromising runs. \\
A first extension overcoming the sequential issue is proposed by \cite{freezethaw}. The instance-dependent Gaussian Process model built from the available learning curves\footnote{In the following, {\em learning curve} denotes the available evidence about a configuration, reporting the performance w.r.t. the number of epochs so far.} is used to decide whether to freeze a run or start another run. Overall, \cite{freezethaw} maintains a basket of runs, typically involving 10 alive (non-frozen) runs and 3 new ones, where the decision is based on the maximum "asymptotic" performance reached on this learning curve according to Expected Improvement. In each round, the GP model is updated and the basket of runs is recomposed.\\
Another approach is that of \cite{Domhan}, with two differences compared to \cite{freezethaw}. Firstly, the domain knowledge is leveraged to select 11 models best reflecting the usual learning curves (ranging from vapor-pressure to Weibull law; see \cite{Domhan} for more detail), and referred to as basic models in the following. The ensemble of these basic models constitute a parametric ensemble modelling space, including the parameters of each model and the weight of each model in the ensemble. Each learning curve is exploited using Bayesian inference to derive a posterior distribution on the ensemble modelling space, best accounting for this learning curve. The exploitation of this posterior distribution via MCMC supports an estimation of the performance that might be reached later on this learning curve, and the confidence thereof. Finally, based on a (user-supplied) confidence level $\delta$, a learning curve is halted whenever the probability that its eventual performance improves on the best-so-far performance is less than $\delta$.


\def\SH{Successive-Halving}
\paragraph{Multi-Armed Bandit.}
Another approach to parallel online optimization and pruning is based on the Multi-Armed Bandit framework, offering rigorous guarantees about the optimal allocation of trials. In \cite{Hyperband}, the problem of hyper-parameter optimization is formulated as a pure exploration adaptive resource allocation problem. The approach builds upon the \SH\ process proposed by \cite{SuccessiveHalving}, which most simply prunes 50\% of the runs with lowest current performance, until a single configuration remains; each run corresponds to a (uniformly sampled) configuration. Naturally, the overall performance of \SH\ critically depends on the initial allocated computational resources. The Hyperband approach \cite{Hyperband} addresses this limitation using a infinitely-many arm bandit approach on the space of number $n$ of configurations to be considered in parallel, times computational time $r$ allocated between two pruning steps, where the instant reward associated to an  $(n,r)$ pair is the best learning performance achieved by \SH$(n,r)$.

\paragraph{Discussion.}
Compared to Hyperband, performance modelling offers two significant advantages. Firstly, it makes it possible to prune an arbitrary number of runs whenever an excellent one is found; in contrast, Hyperband does not allow learning across runs; each trial consider a new iid configuration sample. Secondly, Hyperband involves a fixed discarding rate, determining the fraction of pruned runs in each \SH\ step. In the \DL\ context however, validation curves are very noisy at the beginning, and some hyper-parameters (e.g. when the learning rate decay starts) have a delayed impact, making all learning curves very similar in the early steps. In such contexts, early pruning is mostly random. \\
On the other side, performance modelling does not allow for an efficient pruning in the parallel setting. Typically, whenever several runs are very similar and close from the best-so-far one, the parallel approach proposed by \cite{Domhan} is bound to keep them all. 

Our goal, as said, is to achieve an optimal pruning under the optimality constraint (preserving the optimal performance out of the initial set of configurations). To this aim, the contributions described in next section will focus on how to use the performance model in order to adjust the selection threshold, and how to address  the multiple hypothesis testing issue in a consistent way. 


%% file: Methods.tex
\section{Overview of \PORR}\label{sec:algo}
The presented approach relies on performance modelling and closely follows the approach of
\cite{Domhan}.
The same 11 basic models are used\footnote{As the goal is a minimization one, the model $f_{\theta}(x)$ becomes $\theta' - f_{\theta}(x)$ with $\theta'$ an additional parameter.}. All attempts to reduce the number of models resulted in lesser performances. 
Each learning curve (validation-error($t$), for $t = 1 \ldots $current epoch)
derives a posterior distribution on the ensemble modelling space, using Bayesian inference from the same un-informative prior. This posterior distribution is likewise used by MCMC to derive an estimate of the validation-error for $t' > t$, together with the confidence thereof. 
The overall computational budget is finite, with $t < T$. For simplicity and by abuse of language, we will refer to {\em asymptotic} properties to designate properties that are true at epoch $T$. 

{\bf{Criteria for halting a run.}}
Six criteria are presented below, to make the decision of halting a learning curve (halting the run with no later resuming). These criteria are parameterized by a confidence threshold $\delta \in [0,1]$, as in 
\cite{Domhan}.
Another quantity involved in these criteria is the current best performance noted $y^*(t)$ and {\em the predicted asymptotic result of} the current best learning curve, noted $\hat{y}^*(T)$. 

\begin{enumerate}[(a)]
\item {\bf{Default halt operator}}: a run is halted if the probability that it performs asymptotically better than the current best is less than $\delta$. This is the criterion used by \cite{Domhan}).  

If we trust the confidence intervals and if the validation error is noise-free, this criterion has a probability at least $1-\delta$ not to halt an optimal curve. 
At each given time step, the probability of a mishalt (i.e., halt of the optimal run) is therefore bounded by $\delta$.

\item a run is halted if the probability that it performs asymptotically better than {\em the predicted asymptotic result of} the current best is less than $\delta$. 

Compared to (a), the bound now is the predicted expected performance of the current best, instead of the current result of the current best. But this criterion is risky, as it does not use the confidence interval of the current best; it will therefore not be mentioned any more here, subsumed by next criterion (c).

\item {\bf{Prediction-halt operator}}: a run is halted if the probability that it performs asymptotically better than {\em a conservative estimate of} the predicted asymptotic result of the current best is less than $\delta$. 

This criterion uses an upper bound on the asymptotic performance of the current best run. The decision hence depends on two curve models. There is thus, there is a probability of at least $1-2\delta$ not to halt an optimal run: the probability f mishalt is therefore bounded by $2 \delta$.

A drawback of both (b) and (c) is how they handle currently poor runs that present a steep improvement, whereas the current best is stagnating. In such case, the criterion might be too conservative. This leads to proposing the following criteria (d) and (e), counterparts of criteria (b) and (c) but using the overall best conservative prediction instead of the conservative prediction of the current best.

\item  a run is halted if the probability that it performs asymptotically better than the {\em overall best of all} predicted asymptotic results is less than $\delta$. 

\item  {\bf{ Best-prediction halt operator}}: a run is halted if the probability that it performs asymptotically better than the overall best {\em conservative estimate} of all predicted asymptotic results is less than $\delta$. 

There is however a subtle side-effect with this operator: if the probability of failure is $1-\delta$ for each run, and there are $n$ runs, then the cumulated risk can be $1$!. This is why we propose the following criterion: 

\item  {\bf{Clever-halt operator}}: a run is halted if the probability that it performs asymptotically better than {\em the $k^{th}$ best} predicted asymptotic results is less than $\delta$. We will now discuss the choice of $k$.

\end{enumerate}

Let us assume that there are $n$ competing runs, and that the probability that {\em{one given}} curve modelling fails in providing an upper or lower bound is at most $\delta$ (we trust our models with confidence $\delta$).
Then, the probability that {\em at least one curve} is poorly modeled is less than $1-(1-\delta)^n$ -- and one poor modeling can lead to the failure of methods (d) or (e).

Furthermore, this implies that the probability {\em either} the best asymptotic run or the current best run is poorly modeled is less than $2\delta$: this justifies method (c), but not (b).

Finally, the probability that {\em at least $k$ curves} are poorly modeled can be made less than $\delta$ by choosing $k$ sufficiently large. Here, using the Gaussian approximation of this probability, we want that $k$ satisfies $P({\cal N}(0,1)\geq \frac{k-n\delta}{\sqrt{n \delta (1-\delta)}})\leq \delta$. We select $k$ numerically as the smallest integer satisfying this inequality. As $k$ depends on $n$, it will vary from one context to the next.



%% file: Experiments.tex
\section{Experiments}\label{sec:expe}
This section presents the experimental setting and the experimental methodology
followed to empirically compared the proposed \PORR\ pruning criteria. 
The first experimental results (section \ref{sec:delta}) suggest some simple heuristic improvements (section \ref{sec:hack}). 

\input{Testbeds}

\input{Preliminaries}

\input{Delta}

\input{Hack}

\input{FurtherDelta}

\section{Conclusions}
A first contribution of the presented work is to confirm the relevance of the pruning method proposed by \cite{Domhan}, with computational savings often above 85\% (particularly so for image applications), and applicable at both levels of randomized hyper-parameter optimization, and model selection (the model selection problem itself embedding a hyper-parameter selection problem). Cumulative gains from both levels can decrease the computational cost by more than an order of magnitude as shown on the Cifar and Metacifar experiments. Interestingly, the sensitivity of the approach w.r.t. {\bf the confidence threshold  $\delta$} reveals itself to be low. Set to $0.05$ in \cite{Domhan}, we show that it can be increased up to $0.5$;  this stability is explained from the conservative modelling of the performance in the considered settings. 
Our second contribution is a new and more principled pruning method, slightly but significantly outperforming the former method for all confidence threshold $\delta$, with a  excellent stability with respect to $\delta$  (see Fig.\ref{compa} and supplementary material). The {\bf{novelty:}} of the proposed approach is to refine the halting threshold using predictions, and use a quantile of the predictions (as opposed to, the best prediction), in an adaptive manner. The robustness of this approach relies on its consistent grounding on the multiple hypothesis testing framework. 
A first research perspective 
 is to introduce diversity-based pruning for ensemble methods \cite{ensemblehp}, taking inspiration from the clearing methods in multi-modeal optimization \cite{clearing}. A second perspective will investigate whether the quantile-based proposed approach can make the inference simpler (as opposed to the Metropolis-Hasting method used in \cite{Domhan}).
Finally, combining the proposed approach with mainstream Bayesian Optimization, exploiting both the validation curves and the structure of the hyperparameter domain, would allow to learn accross runs.

%% file: Testbeds.tex
\subsection{Experimental setting}

Two families of large-size problems are considered, within the domain of language modelling and classification.  In the former case, the loss is expressed in terms of perplexity (bits-per-unit, when predicting the next word). In the latter one, the loss is the cross-entropy one unless otherwise stated. Due to the large size of the datasets, only the validation error is considered; the overfitting issue is beyond the scope of the paper. 

Five language modelling tasks are considered. PTB \cite{PTB} aims at language modeling at the bit, byte and word levels. MiniWiki is a subset of the Hutter dataset \cite{hutter} (also referred to as enwik8.zip); the size of the training set is 6\% of the overall size; the modelling task is at the bit and byte level. 
The considered neural architecture involves 3 stacked LSTM with 500 units, batch size 50, 30 unrolling steps, with a budget of 30 epochs. The hyper-parameters (uniformly and independently drawn) are the weight init scale (in $[0.02,1]$), the learning rate (in $[5,100]$), the dropout keep probability  (in $[0.2,1]$), the clipping gradient norm (in $[0.05,1]$). 

Four classification settings are considered, all based on the  Cifar10 dataset \cite{cifar10}, and involving four learning rate adaptation methods, namely  Adagrad, Adam, Gradient, and Momentum. The NN architecture is made of 3 convolutional layers, with filter size 7x7, max-pooling, stride 1, 512 chanels; followed by a convolutional layer with filter size 5x5 and 64 chanels; followed by two fully connected layers with 384 and 192 units respectively. In all cases, the batch size is 64 with a budget of 200 epochs. The hyper-parameters (uniformly and independently drawn)
include the weight init scale (in $[0.001, 100]$), the weight init scale for convolutional layers (in $[0.001,0.1]$), the learning rate (in $[0.00001,10]$), the clipping gradient norm (in $[0.01,10]$), the number of epochs before learning rate decay (in $[12,198]$), the learning rate decay (in $[0.9, 0.999]$) and the dropout keep probability (in $[0.8, 1]$) for non-convolutional layers. 

Specific experiments are considered for investigation:

{\bf{Mini:}} considers the PTB word prediction task, with a small net with 2 layers of 20 units, 30 unrolling steps, learning rate 35, dropout 0.5, gradient clipping norm 0.143, 30 epochs, cell clipping \cite{cellclip} between $1$ and $10000$.

{\bf{Maxi:}} only differs from Mini as it considers more runs in parallel. 

{\bf{Coupled:}} considers the PTB bytes or word prediction task, with a larger net involving 2 LSTM with 650 units, optimized by stochastic gradient descent, optionally with coupled input and forget gates, other values as in Mini.

{\bf{MetaCifar:}} operates on four experiments, each one considering 22 runs with a different learning rate adaptation method (Adagrad, Adam, Gradient, and Momentum). 

{\bf{Metanormalization:}} considers 32 runs with different variants of normalization for language modeling with three different toy sequences: (i) rote learning of sequences ``.M'' made of  repetition of identical words; (ii) sequences `AN'' repeating identical repetitions of same length words made of a same letter (but with possibly different lengths for different sequences); (iii) sequences of the form ``anbn'' (aaabbb aaaaaabbbbb\dots) with varying $n$.

%% file: Preliminaries.tex
\subsection{Experimental methodology}
\begin{figure}[htbp]
\center
\includegraphics[width=.49\textwidth]{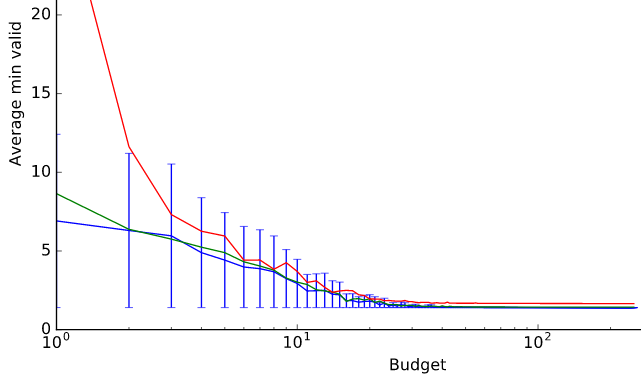}
\includegraphics[width=.49\textwidth]{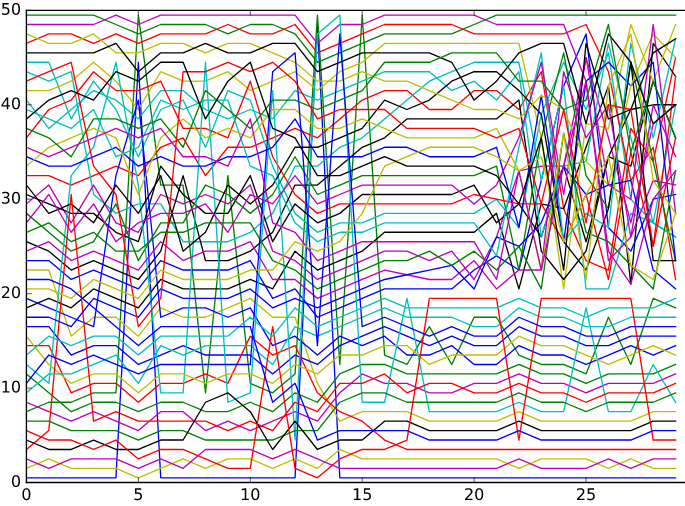}
\caption{Impact of the budget on the best validation performance on the PTB modelling task at the bit level, considering 50 runs. \label{budget} Left: Best validation performance vs the computational budget. The three curves respectively correspond to an allowed number of epochs set to 30, 15, and 7. The confidence intervals correspond to the ``30 epochs'' curve. 
Right: variability of the ranks of different runs (PTB bits, 50 runs; best seen in color).}
\label{fig:budget}
\end{figure}

\begin{figure}
\centering
\includegraphics[width=.48\textwidth]{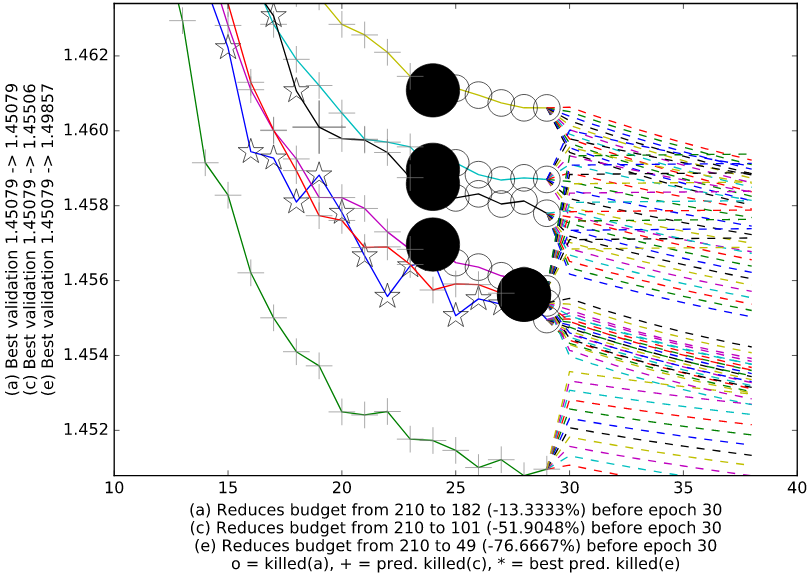}
\includegraphics[width=.48\textwidth]{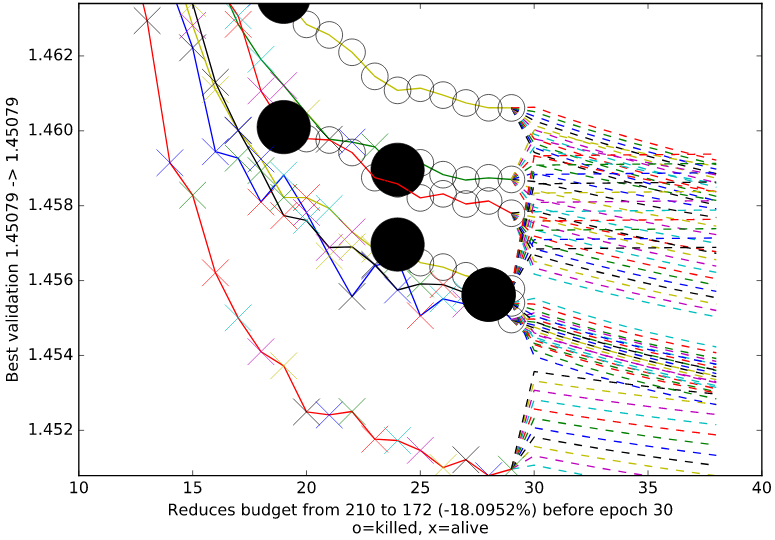}
\caption{{\sc Mini} case 
with $\delta=0.5$; a black ball indicates that the run is halted. Left: methods  (a), (c) and (e); legend $+$ indicates that the best run was killed by (c); legend $\ast$ indicates that the second best run was killed only by (e). 
Right: method (f), halting non-optimal curves earlier than (a).
}
\label{illus}\label{fig:illus}
\end{figure}

In the remainder of the paper, a failure (FAIL) stands for halting the best run. 
The main components for the \PORR\ decisions are: i) the confidence threshold $\delta$ needed to prune a run; ii) the comparison threshold: the pruning decision is taken if the predictive performance of a run is below the comparison threshold with confidence at least 1 - $\delta$, and iii) the overall computational budget. 
The sensitivity w.r.t. the confidence value is discussed in section  \ref{sec:delta}. 
The impact of the overall computational budget (here, the number of simultaneous runs) is dramatic, as illustrated on Fig. \ref{fig:budget} in the case of the 
PTB modelling task at the bit level. 50 runs are launched in parallel, with an allowed number of epochs set to 30, 15, and 7. Empirically, circa 30 independent parallel runs are required to eventually deliver a "reasonably optimal" performance. Note that this experimental setting is a typical one, with a very significant computational cost; hence the need for the present work. \\
Additionally, Fig. \ref{fig:budget} empirically demonstrates that the early ranks of the runs can be very misleading.
The experimental observations on a given problem are illustrated on the Mini case (Fig. \ref{fig:illus}) for $\delta = .5$. 
In the following, the  performance of a pruning criterion on a given problem is assessed depending on whether it failed and halted the best run (FAIL); otherwise, it reports the computational savings.

%% file: Delta.tex
\subsection{Pruning with confidence $\delta = .5$} \label{sec:delta}
Table \ref{maintable} reports the results of criteria (a) (the baseline \cite{Domhan}); (c), (e) and (f) on all experimental cases, for $\delta = .5$. 

Unexpectedly, although the confidence is very low (which might imply at first sight that the probability of FAIL is ca 50\%), these results are very good. A tentative interpretation for this fact is that, although the confidence is low, the comparison threshold is set to the best validation error so far, which is a quite conservative threshold. With a low confidence $\delta$, more aggressive comparison thresholds entail failures. Criterion (c), considering the predicted error threshold, fails. Criterion (e), considering the best predicted validation error, fails even more often. Overall, the baseline method (a) and method (e) do not fail. The computational savings are such that method (e) wins over (a) in 5 experimental cases, and (a) wins over (e) in 1 experimental case. 

\begin{table}[t]
\center
{\scriptsize{\begin{tabular}{|@{}m{2cm}@{}|@{}m{2cm}@{}|@{}m{2cm}@{}|@{}m{2cm}@{}|@{}m{2cm}@{}|@{}m{2cm}@{}|}
\hline
 Testbed &
 \begin{minipage}{2cm}Budget (number of runs)\end{minipage} &
 \begin{minipage}{2cm}Computational cost saved up by method (a) (equal to Domhan et al) \end{minipage}&
 \begin{minipage}{2cm}C.c. saved up by method (c): prediction-halt operator. \end{minipage}&
 \begin{minipage}{2cm}C.c. saved up by method (e): best-prediction-halt operator. \end{minipage}&
 \begin{minipage}{2cm}C.c. saved up by method ``Clever-halt''\end{minipage} \\
\hline
 Cifar-adagrad  &
 22 &
 \bf -89.6\% &
 -93.5\% FAIL by 1.083 $\rightarrow$ 1.164 &
 -96.5\% FAIL by 1.083 $\rightarrow$ 1.164 &
 \bf -89.6\% \\
\hline
 Cifar-adam &
 22 &
 \bf -89.2\% &
 -91.5\% FAIL by 0.80 $\rightarrow$ 0.92 &
 -96.5\% FAIL by 0.80 $\rightarrow$ 0.94  &
 \bf -89.2\% \\
\hline
 Cifar-gradient &
 22 &
 \bf -87.2\% &
 \bf -87.2\% &
 FAIL by 1.18 $\rightarrow$ 1.95 &
 \bf -87.2\% \\
\hline
 Cifar-momentum &
 22 &
 \bf -85.5\% &
 -96.2\% FAIL by 1.03 $\rightarrow$ 1.45 &
 -96.2\% FAIL by 1.03 $\rightarrow$ 1.45 &
 \bf -85.5\% \\
\hline
 Miniwiki bits &
 250 &
 \bf -69.6\% &
 -74.1\% FAIL by 2.01 $\rightarrow$ 2.06 &
 -76.6\% FAIL by 2.01 $\rightarrow$ 2.52 &
 \bf -69.6\% \\
\hline
 Miniwiki bytes &
 250 &
 \bf -47.6\% &
 -47.6\% &
 -73.2\% FAIL by 1.86 $\rightarrow$ 1.91 &
 -46.9\% \\
\hline
 PTB bits &
 250 &
 -68.4\% &
 -74.5\% FAIL by 1.40 $\rightarrow$ 1.56 &
 -76.7\% FAIL by 1.40 $\rightarrow$ 1.66 &
 \bf -69.7\% \\
\hline
 PTB bytes &
 250 &
 \bf -65.7\% &
 FAIL by 1.298 $\rightarrow$ 1.317 &
 FAIL by 1.298 $\rightarrow$ 1.419 &
 \bf -65.7\% \\
\hline
 PTB words &
 250 &
 -72.1\% &
 \bf -75.1\% &
 -76.6\% FAIL by 1.18 $\rightarrow$ 1.24 &
 -72.1\% \\
\hline
 Miniwiki bits &
 50 &
 -72.1\% &
 \bf -73.0\% &
 -76.3\% FAIL by 2.02 $\rightarrow$ 2.54 &
 -72.1\% \\
\hline
 Miniwiki bytes &
 50 &
 \bf -61.5\% &
 -72.6\% FAIL by 1.88 $\rightarrow$ 2.03 &
 -76.7\% FAIL by 1.88 $\rightarrow$ 2.19 &
 \bf -61.5\% \\
\hline
 PTB bits &
 50 &
 \bf -72.1\% &
 -74.5\% FAIL by 1.41 $\rightarrow$ 1.46 &
 -76.7\% FAIL by 1.41 $\rightarrow$ 1.70 &
 \bf -72.1\% \\
\hline
 PTB bytes &
 50 &
 \bf -62.1\% &
 -72.9\% FAIL by 1.31 $\rightarrow$ 1.40 &
 -76.7\% FAIL by 1.31 $\rightarrow$ 1.47 &
 \bf -62.1\% \\
\hline
 PTB words &
 50 &
 \bf -65.1\% &
 -70.5\% FAIL by 1.20 $\rightarrow$ 1.22 &
 -76.5\% FAIL by 1.20 $\rightarrow$ 1.28 &
 \bf -65.1\% \\
\hline
 Mini &
 7 &
 -13.3\% &
 -55.2\% FAIL by 1.455 $\rightarrow$ 1.149 &
 -76.7\% FAIL by 1.455 $\rightarrow$ 1.499 &
 \bf -18.1\% \\
\hline
 Uncoupled bytes &
 5 &
 -5.33\% &
 -70\% FAIL by 1.402 $\rightarrow$ 1.195 &
 -76.7\% FAIL by 1.402 $\rightarrow$ 1.625 &
 \bf -10.7\% \\
\hline
 Coupled bytes &
 5 &
 -2.7\% &
 -73.3\% FAIL by 1.155 $\rightarrow$ 1.176 &
 -73.3\% FAIL by 1.155 $\rightarrow$ 1.176 &
 \bf -6.7\% \\
\hline
 Uncoupled words &
 5 &
 \bf -5.33\% &
 -49.3\% FAIL by 1.470 $\rightarrow$ 1.471 &
 -76.7\% FAIL by 1.470 $\rightarrow$ 1.767 &
 \bf -5.33\% \\
\hline
 Coupled words &
 5 &
 -2.67\% &
 -74\% FAIL by 1.16 $\rightarrow$ 1.23 &
 -75.3\% FAIL by 1.16 $\rightarrow$ 1.23 &
 \bf -9.33\% \\
\hline
 Maxi &
 77 &
 -4.91\% &
 -17.6\% &
 -21.93\% FAIL by 1.482 $\rightarrow$ 1.490 &
 \bf -5.48\% \\
\hline
MetaCifar   &
4x22 & 
  \bf -42.5\%  &
    -62.7\% FAIL by 1.03 $\rightarrow$ 1.10 &
      -75.4\% FAIL by 1.03 $\rightarrow$ 1.10 &
\bf        -42.5\% \\
\hline
MetaNorm .N & 32x1 & \bf -53.5\% & -56.5\% FAIL by 0.93 $\rightarrow$ 0.98 & -56.9\% FAIL by 0.93 $\rightarrow$ 0.98 & \bf -53.5\%\\ 
MetaNorm AN & 32x1 & -68.9\% & \bf -69.6\% & \bf -69.6 \% & -68.9\% \\
MetaNorm anbn & 32x1 & \bf -72.6\% & -73.5\% FAIL by $3e-4$ & -73.5\% FAIL by $3e-4$ & \bf 72.6\%  \\
\hline
 & & & & &  No failure, best average performance. \\
\hline
\end{tabular}}}
\caption{\label{maintable}Comparative results of pruning criteria (a) \cite{Domhan}, (c), (e) and (f) (respectively in columns 3, 4, 5 and 6), with confidence $\delta=0.5$.}
\end{table}

%
%

%% file: Hack.tex
\subsection{Conservative pruning rules}\label{sec:hack}
A natural question raised from the empirical results (Table \ref{maintable}) is whether undesirable failures could be prevented using simple heuristic conservative rules, such as: i) never prune the current best; ii) discard all predictions of a negative loss; iii) discard predictions with correlation data/observation less than 0.5. 
Accordingly, the experiments are repeated by enriching all pruning criteria with 
the conservative rules (Table \ref{withhacks}). Although these simple rules do save some failures for methods (c) and (e), the overall conclusions remain the same as from Table \ref{maintable}: methods (a) and (f) are the only safe ones, with (e) slightly outperforming (a) in terms of computational savings. As expected, the overall gain is eroded as the aggressive (c) and (e) methods do no longer fail on 4 out of the 15 problems.  

\begin{table}[t]
\center
{\scriptsize{\begin{tabular}{|@{}m{2cm}@{}|@{}m{2cm}@{}|@{}m{2cm}@{}|@{}m{2cm}@{}|@{}m{2cm}@{}|@{}m{2cm}@{}|}
\hline
 Testbed &
 Budget (number of runs) &
 Computational cost saved up by method (a) (equal to Domhan et al) &
 C.c. saved up by method (c): prediction-halt operator. &
 C.c. saved up by method (e): best-prediction-halt operator. &
 C.c. saved up by method ``Clever-halt'' \\
\hline
 Cifar-adagrad  &
 22 &
 \bf -56.5\% &
 \bf -56.5\% &
 \bf -56.5\% &
 \bf -56.5\% \\
\hline
 Cifar-adam &
 22 &
 \bf -88.1\% &
 -94.1\% FAIL by 0.797 $\rightarrow$ 0.923 &
 -95.1\% FAIL by 0.797 $\rightarrow$ 0.939 &
\bf  -88.1\% \\
\hline
 Cifar-gradient &
 22 &
 \bf -83.5\% &
 \bf -83.5\% &
 \bf -83.5\% &
 \bf -83.5\% \\
\hline
 Cifar-momentum &
 22 &
 \bf -82.5\% &
 -92.5\% FAIL by 1.034 $\rightarrow$ 1.451 &
 -92.9\% FAIL by 1.034 $\rightarrow$ 1.451 &
 \bf -82.5\% \\
\hline
 Miniwiki bits &
 250 &
 \bf -48.6\% &
 -54.7\% FAIL by 2.011 $\rightarrow$ 2.035 &
 -54.7\% FAIL by 2.011 $\rightarrow$ 2.035 &
 \bf -48.6\% \\
\hline
 Miniwiki bytes &
 250 &
 \bf -13.8\% &
 -13.8\% &
 -14.1\% &
 \bf -13.8\% \\
\hline
 PTB bits &
 250 &
 \bf -44.0\% &
 -44.8\% FAIL by 6e-4 &
 -44.8\% FAIL by 6e-4 &
 \bf -44.0\% \\
\hline
 PTB bytes &
 250 &
 -20.6\% &
 \bf -21.2\% &
 \bf -21.2\% &
 -20.6\% \\
\hline
 PTB words &
 250 &
 -66\% &
 \bf -69.7\% &
 \bf -69.7\% &
 -66\% \\
\hline
 Miniwiki bits &
 50 &
 -41.8\% &
 \bf -42.7\% &
 \bf -42.7\% &
 -41.8\% \\
\hline
 Miniwiki bytes &
 50 &
 -43.8\% &
 \bf -49.9\% &
 \bf -49.9\% &
 -43.8\% \\
\hline
 PTB bits &
 50 &
 \bf -50.1\% &
 -50.2\% FAIL by 2e-4 &
 -50.2\% FAIL by 2e-4 &
 \bf -50.1\% \\
\hline
 PTB bytes &
 50 &
 \bf -22\% &
 \bf -22\% &
 \bf -22\% &
 \bf -22\% \\
\hline
 PTB words &
 50 &
 \bf -60.5\% &
 -68.7\% FAIL by 1.201 $\rightarrow$ 2.215 &
 -73.3\% FAIL by 1.201 $\rightarrow$ 2.215 &
 \bf -60.5\% \\
\hline
 Mini &
 7 &
 -13.3\% &
 -51.9\% FAIL by 1.451 $\rightarrow$ 1.455 &
 -76.7\% FAIL by 1.451 $\rightarrow$ 1.499 &
 \bf -18.1\% \\
 \hline 
\end{tabular}}}
\caption{\label{withhacks}Comparative results under same conditions as in Table \ref{maintable}, where each pruning criterion is enriched with three simple conservative rules.}
\end{table}

%% file: FurtherDelta.tex
\subsection{Pruning with low confidence $\delta$ and conservative rules} \label{sec:furtherdelta}
Assuming that the conservative heuristic rules will prevent aggressive pruning criteria from most failures, a question is whether better savings can be obtained by lowering the confidence threshold. As shown in Table \ref{zerozeroone} (supplementary material), a lower $\delta = .01$ does significantly improve the results for criterion (c) (though dominated by the results of criterion (f) for $\delta = .5$). For criterion (e) however, failures are still observed; the interpretation for this fact (in agreement with the multiple hypothesis testing framework indeed) is that increasing the number of tests is a strong factor of failure. 
Interestingly, the overall results are globally worse than for $\delta=0.5$.

Other confidence levels ($\delta=0.1$, $\delta=0.3$, $\delta=0.05$, $\delta=0.01$) have been considered (results in supplementary material). Fig. \ref{compa}
 graphically displays the performance of the pruning criterion (f) compared to the baseline (a) \cite{Domhan}.
\begin{figure}
\centering
\includegraphics[width=.48\linewidth]{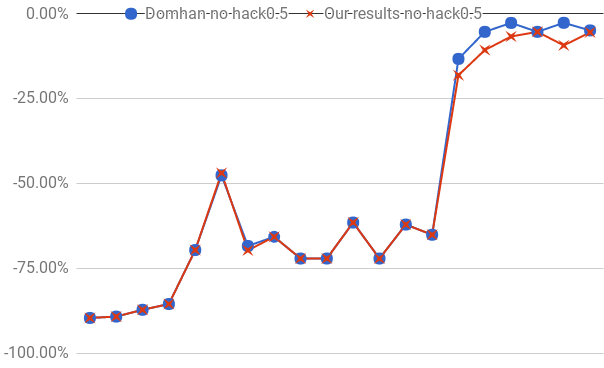}
\includegraphics[width=.48\linewidth]{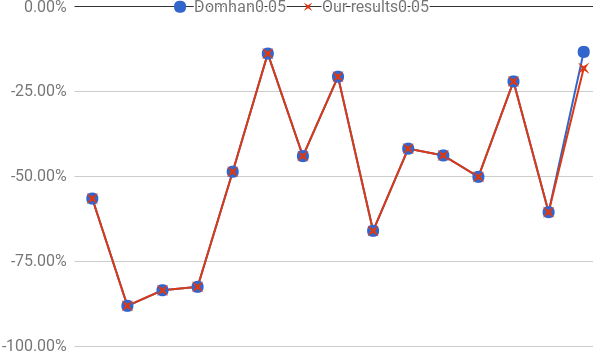}\\
\includegraphics[width=.48\linewidth]{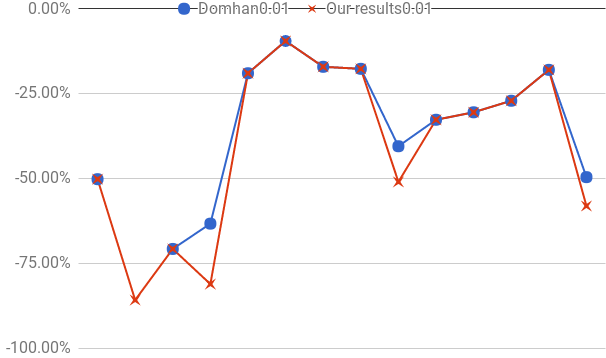}
\includegraphics[width=.48\linewidth]{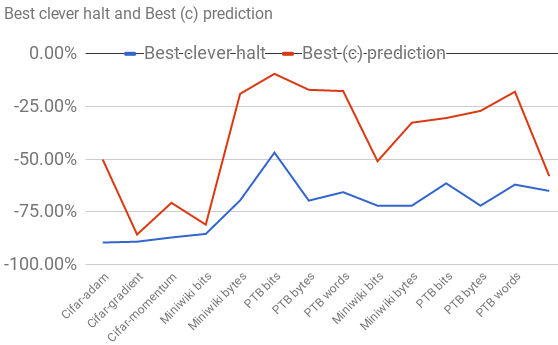}\\
\caption{Top left, top right, bottom left: comparison between the baseline (a) and the contributed pruning criterion (f); for each value of the confidence parameter $\delta$, (f) improves on (a) by a small but significant amount.\label{compa}
Bottom right: comparison between the best results (with no FAIL) for $\delta=0.01$, and (f) criterion with $\delta=0.5$.}
\end{figure}

%% file: ms.bbl
\begin{thebibliography}{10}

\bibitem{AutoML15-NIPS}
{\em Challenges in Machine Learning}, 2015.

\bibitem{Bardenet13}
R.~Bardenet, M.~Brendel, B.~K{\'e}gl, and M.~Sebag.
\newblock Collaborative hyperparameter tuning.
\newblock In {\em Proceedings of the 30th International Conference on Machine
  Learning (ICML)}, volume~28, pages 199--207, 2013.

\bibitem{Bardenet11}
J.~Bergstra, R.~Bardenet, Y.~Bengio, and B.~K{\'e}gl.
\newblock Algorithms for hyper-parameter optimization.
\newblock In P.~B. F. P. K.~W. J.~Shawe-Taylor, R.S.~Zemel, editor, {\em
  Proceedings of the 25th Annual Conference on Neural Information Processing
  Systems (NIPS)}, volume~24 of {\em Advances in Neural Information Processing
  Systems}, Granada, Spain, 2011.

\bibitem{Bradzil00}
P.~Brazdil and C.~Soares.
\newblock A comparison of ranking methods for classification algorithm
  selection.
\newblock In R.~L. de~M{\'a}ntaras and E.~Plaza, editors, {\em the 11th
  European Conference on Machine Learning (ECML)}, volume 1810 of {\em LNCS},
  pages 63--74. Springer, 2000.

\bibitem{cellclip}
W.~Chan and I.~Lane.
\newblock Deep recurrent neural networks for acoustic modelling.
\newblock {\em CoRR}, abs/1504.01482, 2015.

\bibitem{Deng}
L.~Deng, G.~E. Hinton, and B.~Kingsbury.
\newblock New types of deep neural network learning for speech recognition and
  related applications: an overview.
\newblock In {\em {ICASSP}}, pages 8599--8603. {IEEE}, 2013.

\bibitem{Domhan}
T.~Domhan, J.~T. Springenberg, and F.~Hutter.
\newblock Speeding up automatic hyperparameter optimization of deep neural
  networks by extrapolation of learning curves.
\newblock In {\em IJCAI}, pages 3460--3468. AAAI Press, 2015.

\bibitem{Donahue}
J.~Donahue, Y.~Jia, O.~Vinyals, J.~Hoffman, N.~Zhang, E.~Tzeng, and T.~Darrell.
\newblock Decaf: {A} deep convolutional activation feature for generic visual
  recognition.
\newblock In {\em ICML}, volume~32 of {\em {JMLR} Workshop and Conference
  Proceedings}, pages 647--655. JMLR.org, 2014.

\bibitem{SuccessiveHalving}
A.~Gretton and C.~C. Robert, editors.
\newblock {\em Non-stochastic Best Arm Identification and Hyperparameter
  Optimization}, volume~51 of {\em {JMLR} Workshop and Conference Proceedings}.
  JMLR.org, 2016.

\bibitem{AutoML16-ICML}
F.~Hutter, L.~Kothoff, and J.~Vanshoren.
\newblock Automl 2016, 2016.

\bibitem{hutter}
M.~Hutter.
\newblock 50000 euros prize for compressing human knowledge, 2006.

\bibitem{hord}
I.~Ilievski, T.~Akhtar, J.~Feng, and C.~A. Shoemaker.
\newblock Hyperparameter optimization of deep neural networks using
  non-probabilistic {RBF} surrogate model.
\newblock {\em CoRR}, abs/1607.08316, 2016.

\bibitem{cifar10}
A.~Krizhevsky, V.~Nair, and G.~Hinton.
\newblock Cifar-10 (canadian institute for advanced research).

\bibitem{KrizhevskySH12}
A.~Krizhevsky, I.~Sutskever, and G.~E. Hinton.
\newblock Imagenet classification with deep convolutional neural networks.
\newblock In P.~L. Bartlett, F.~C.~N. Pereira, C.~J.~C. Burges, L.~Bottou, and
  K.~Q. Weinberger, editors, {\em NIPS 12}, pages 1106--1114, 2012.

\bibitem{ensemblehp}
J.~Levesque, C.~Gagn{\'{e}}, and R.~Sabourin.
\newblock Bayesian hyperparameter optimization for ensemble learning.
\newblock {\em CoRR}, abs/1605.06394, 2016.

\bibitem{Hyperband}
L.~Li, K.~G. Jamieson, G.~DeSalvo, A.~Rostamizadeh, and A.~Talwalkar.
\newblock Efficient hyperparameter optimization and infinitely many armed
  bandits.
\newblock {\em CoRR}, abs/1603.06560, 2016.

\bibitem{PTB}
M.~P. Marcus, M.~A. Marcinkiewicz, and B.~Santorini.
\newblock Building a large annotated corpus of english: The penn treebank.
\newblock {\em Comput. Linguist.}, 19(2):313--330, June 1993.

\bibitem{Atari}
V.~Mnih and al.
\newblock Human-level control through deep reinforcement learning.
\newblock {\em Nature}, 518(7540):529--533, 2015.

\bibitem{Bayesian-Mockus}
J.~Mockus.
\newblock Bayesian heuristic approach to global optimization and examples.
\newblock {\em J. Global Optimization}, 22(1-4):191--203, 2002.

\bibitem{Kalousis14}
P.~Nguyen, M.~Hilario, and A.~Kalousis.
\newblock Using meta-mining to support data mining workflow planning and
  optimization.
\newblock {\em J. Artif. Intell. Res. {(JAIR)}}, 51:605--644, 2014.

\bibitem{clearing}
A.~P\'etrowski.
\newblock A clearing procedure as a niching method for genetic algorithms.
\newblock In {\em International Conference on Evolutionary Computation}, pages
  798--803, 1996.

\bibitem{rice1976algorithm}
J.~Rice.
\newblock The algorithm selection problem.
\newblock {\em Advances in computers}, 15:65--118, 1976.

\bibitem{AlphaGo}
D.~Silver and al.
\newblock Mastering the game of go with deep neural networks and tree search.
\newblock {\em Nature}, 529(7587):484--489, 2016.

\bibitem{SnoekLA12}
J.~Snoek, H.~Larochelle, and R.~P. Adams.
\newblock Practical bayesian optimization of machine learning algorithms.
\newblock In P.~L. Bartlett, F.~C.~N. Pereira, C.~J.~C. Burges, L.~Bottou, and
  K.~Q. Weinberger, editors, {\em NIPS12}, pages 2960--2968, 2012.

\bibitem{Pfahringer13}
Q.~Sun and B.~Pfahringer.
\newblock Pairwise meta-rules for better meta-learning-based algorithm ranking.
\newblock {\em Machine Learning}, 93(1):141--161, 2013.

\bibitem{freezethaw}
K.~Swersky, J.~Snoek, and R.~P. Adams.
\newblock Freeze-thaw bayesian optimization.
\newblock {\em CoRR}, abs/1406.3896, 2014.

\bibitem{thornton2013AutoWeka}
C.~Thornton, F.~Hutter, H.~H. Hoos, and K.~Leyton-Brown.
\newblock {Auto-WEKA}: Combined selection and hyperparameter optimization of
  classification algorithms.
\newblock In {\em Proceedings of the 19th ACM SIGKDD International Conference
  on Knowledge Discovery and Data Mining (KDD)}, pages 847--855. ACM, 2013.

\bibitem{AutoWeka}
C.~Thornton, F.~Hutter, H.~H. Hoos, and K.~Leyton{-}Brown.
\newblock Auto-weka: combined selection and hyperparameter optimization of
  classification algorithms.
\newblock In I.~S. Dhillon, Y.~Koren, R.~Ghani, T.~E. Senator, P.~Bradley,
  R.~Parekh, J.~He, R.~L. Grossman, and R.~Uthurusamy, editors, {\em The 19th
  {ACM} {SIGKDD} International Conference on Knowledge Discovery and Data
  Mining, (KDD)}, pages 847--855, 2013.

\end{thebibliography}
